%% file: acl2019.tex
\documentclass[11pt,a4paper]{article}
\usepackage[hyperref]{acl2019}
\usepackage{times}
\usepackage{latexsym}
\usepackage{amsmath}
\usepackage{mdwlist}
\usepackage{dirtytalk}
\usepackage[ruled]{algorithm2e}
\usepackage{textcomp}

\newcommand{\abr}[1]{\textsc{#1}}
\newcommand{\g}{\, | \,}

\usepackage{url}
\usepackage[colorinlistoftodos]{todonotes}
\aclfinalcopy 
\def\aclpaperid{1757} 

\title{Why Didn't You Listen to Me? Comparing User Control of Human-in-the-Loop Topic Models}

\author{%
    Varun Kumar
    \thanks{Work performed at University of Maryland, College Park}  \\
  Amazon Alexa \\
  Cambridge, MA \\
   \texttt{varunk@cs.umd.edu} \\
   \And
   Alison Smith-Renner \\
  University of Maryland \\
  College Park, MD \\
  \texttt{amsmit@cs.umd.edu} \\
   \AND
   Leah Findlater \\
   University of Washington \\
   Seattle, Washington \\
   \texttt{leahkf@uw.edu} \\
   \And
   Kevin Seppi \\
   Brigham Young University \\
   Provo, UT \\
   \texttt{kseppi@gmail.com} \\
   \And
   Jordan Boyd-Graber \\
   University of Maryland \\
   College Park, MD \\
   \texttt{jordanbg@umiacs.edu} \\
}

\date{}

\begin{document}
\maketitle

\input{sections/00-abstract}
\input{sections/10-introduction}
\input{sections/20-method}

\input{sections/30-results}
\input{sections/40-conclusion}
\section*{Acknowledgements}

This work was supported by the collaborative NSF Grant
\abr{iis}-1409287 (\abr{umd}) and \abr{iis}-1409739 (\abr{byu}). Boyd-Graber is
also supported by \abr{nsf} grant \abr{iis}-1822494 and \abr{iis}-1748663.
Any opinions, findings, conclusions, or recommendations expressed here are those of the authors and do not necessarily
reflect the view of the sponsor.

\bibliography{bib/journal-full,bib/jbg,bib/itm}
\bibliographystyle{acl_natbib}

\appendix
\input{sections/appendix}
\end{document}

%% file: sections/00-abstract.tex
\begin{abstract}
To address the lack of comparative evaluation of Human-in-the-Loop Topic Modeling (\abr{hltm}) systems, we implement and evaluate three contrasting \abr{hltm} modeling approaches using simulation experiments. These approaches extend previously proposed frameworks, including constraints and informed prior-based methods. Users should have a sense of control in \abr{hltm} systems, so we propose a \emph{control} metric to measure whether refinement operations' results match users' expectations. Informed prior-based methods provide better control than constraints, but constraints yield higher quality topics. 
\end{abstract}

%% file: sections/10-introduction.tex
\section{Human-in-the-Loop Topic Modeling}
Topic models help explore large, unstructured text corpora by automatically discovering the topics discussed in the documents~\cite{blei2003latent}. However, generated topic models are not perfect; they may contain incoherent or loosely connected topics~\cite{chang2009reading,mimno2011optimizing,boyd2014care}.

Human-in-the-Loop Topic Modeling (\abr{hltm}) addresses these issues by incorporating human knowledge into the modeling process. Existing \abr{hltm} systems expose topic models as their topic words and documents, and users provide feedback to improve the models using varied refinement operations, such as adding words to topics, merging topics, or removing documents~\cite{smith2018closing,Wang2019InteractiveTM}. Systems also vary in how they incorporate feedback, such as ``must-link'' and ``cannot-link'' constraints~\cite{andrzejewski2009incorporating, hu2014interactive}, informed priors~\cite{smith2018closing}, or document labels~\cite{yang2015efficient}. However, evaluations of these systems are either not comparative~\cite{choo2013utopian, lee2017human} or compare against non-interactive models~\cite{hoque2015convisit, hu2014interactive} or for only a limited set of refinements~\cite{yang2015efficient, xie2015incorporating}. 
Evaluations are thus silent on which \abr{hltm} system best supports users in improving topic models: they ignore whether refinements are applied correctly or how they compare with other approaches. Moreover, comparative evaluations can be difficult because existing \abr{hltm} systems support diverse refinement operations with little overlap. 

To address these issues, we implement three \abr{hltm} systems that differ in the techniques for incorporating prior knowledge (informed priors vs. constraints) and for inference (Gibbs sampling vs. variational \abr{em}), but that all support seven refinement operations preferred by end users~\cite{lee2017human, musialek2016using}. We compare these systems through experiments simulating random and ``good'' user behavior. 
The two Gibbs sampling-based systems extend prior work~\cite{yang2015efficient, smith2018closing}, but to our knowledge, the combination of informed priors and variational inference in an \abr{hltm} system is new. Additionally, while \citeauthor{yang2015efficient} incorporate word correlation knowledge and document label knowledge into topic models, this paper extends their modeling approach with the implementation of seven new user refinements. 

\clearpage

We also introduce metrics to assess the degree to which \abr{hltm} systems listen to users---\emph{user control}---a key user interface design principle for human-in-the-loop systems~\cite{amershi2014power, duchi2017}. In general, informed priors provide more control while constraints produce higher quality topics.

This paper provides three contributions: (1) implementation of an \abr{hltm} system using informed priors and variational inference, (2) experimental comparison of three \abr{hltm} systems, and (3) metrics to evaluate user control in \abr{hltm} systems. 

%% file: sections/20-method.tex
\section{Human Feedback and \abr{LDA}}

We briefly describe Latent Dirichlet Allocation~\cite[LDA]{blei2003latent} and outline the experimental conditions and our implementation.

\subsection{\abr{lda} Inference}
\abr{lda} is generative, modeling documents as mixtures of $k$ topics where each topic is a multinomial distribution, $\phi_{z}$, over the vocabulary, $V$. Each document~$d$ is an admixture of topics~$\theta_{d}$. Each word indexed by~$i$ in document~$d$ is generated by first sampling a topic assignment~$z_{d,i}$ from $\theta_{d}$ and then sampling a word from the corresponding topic~$\phi_{z_{i}}$. 

Collapsed Gibbs sampling \cite{griffiths2004finding} and variational Expectation-Maximization \cite[\abr{em}]{blei2003latent} are two popular inference methods to compute the posterior, $p(z, \phi, \theta \g w, \alpha, \beta)$. Gibbs sampling iteratively samples a topic assignment, $z_{d,i}=t$ given an observed token~$w_{d,i}$ in document $d$ and other topic assignments, $z_{-d,n}$, with probability
\begin{equation}
P(z_{d,i}=t \g z_{-d,n}, w) \propto (n_{d,t}+\alpha) \frac{n_{w,t}+\beta}{n_{t}+V\beta}
\end{equation}
Here, $n_{d,t}$ is the count topic~$t$ is in document~$d$, $n_{w,t}$ is the count of token~$w$ in topic~$t$, and $n_t$ is the marginal count of tokens assigned to topic~$t$. Alternatively, variational \abr{em} approximates the posterior using a tractable family of distributions by first defining a mean field variational distribution 
\begin{align}
 q(z, \phi, \theta \g \lambda, \gamma, \pi) =  \prod_{k=1}^{K} & q(\phi_k \g \lambda_k) \prod_{d=1}^{D} q(\theta_d \g \gamma_d) \notag \\ & \prod_{n=1}^{N_d} q(z_{dn} \g \pi_{dn})
\end{align}
where $\gamma_d$, $\pi_d$ are local parameters of the distribution~$q$ for document~$d$, and $\lambda$ is a global parameter. Inference minimizes the \abr{kl} divergence between the variational distribution and true posterior.  While there are many \abr{lda} variants for specific applications~\cite{boyd-graber-17}, we focus on models that interactively refine initial topic clustering. 

\subsection{\abr{hltm} Modeling Approaches}
To investigate adherence to user feedback and topic quality improvements, we compare \abr{hltm} systems, based on three modeling approaches. Each of these approaches incorporate user feedback by first \textit{forgetting} what the model learned before, by unassigning words from topics~\cite{hu2014interactive}, and then \textit{injecting} new information based on user feedback into the model. 

We compare two existing techniques for \textit{injecting} new information: (1) asymmetric priors (or informed priors), which are used extensively for injecting knowledge into topic models~\cite{fan2017prior,zhai2012mr,pleple2013interactive,smith2018closing,Wang2019InteractiveTM} by modifying Dirichlet parameters, $\alpha$ and $\beta$, and (2) constraints~\cite{yang2015efficient}, in which knowledge source $m$ is incorporated as a potential function $f_m(z, m, d)$ of the hidden topic~$z$ of word type $w$ in document $d$. While other frameworks exist~\cite{foulds2015latent, andrzejewski2009incorporating,hu2014interactive,xie2015incorporating,roberts2014structural}, we focus on informed priors and constraints, as these are flexible to support the refinement operations preferred by users and reasonably fast enough to support ``rapid interaction cycles'' required for effective interactive systems~\cite{amershi2014power}. 

We also compare two inference techniques for topic models (1) Gibbs sampling and (2) variational \abr{em} inference. Because \abr{hltm} requires \textit{forgetting} existing topic assignments~\cite{hu2014interactive}, we use two different methods to forget existing topic assignments. In Gibbs sampling, information is forgotten by adjusting topic-word assignments, $z_i$. In variational \abr{em}, $\lambda_{t,w}$ encodes how closely the word $w$ is related to topic~$t$. In the \emph{E-step}, the model assigns latent topics based on the current value of $\lambda$, and in the \emph{M-step}, the model updates $\lambda$ using the current topic assignments. Because the model relies on a fixed~$\lambda$ for topic assignment, information for a word $w$ in a topic~$t$ can be forgotten by resetting $\lambda_{t,w}$ to the prior~$\beta_{t,w}$. Together, these injection and inference techniques result in three \abr{hltm} modeling approaches:


\paragraph{Informed priors using Gibbs sampling (\textit{info-gibbs})} forgets topic-word assignments $z_i$ and injects new information by modifying Dirichlet parameters, $\alpha$ and $\beta$. \citet{smith2018closing} implement seven refinements for this approach. We extend their work with a create topic refinement.

 \paragraph{Informed priors using variational inference (\textit{info-vb})} forgets topic-word assignments for a word $w$ in topic $t$ by resetting the value of $\lambda_{t,w}$. This approach manipulates priors, $\alpha$ and $\beta$, to incorporate new knowledge like \textit{info-gibbs}. We define and implement seven \textit{user-preferred} refinement operations for this approach.
 
\paragraph{Constraints using Gibbs sampling (\textit{const-gibbs})} forgets topic assignments like in \textit{info-gibbs}, but instead of prior manipulation, injects new information into the model using potential functions, $f_m(z, m, d)$ 
\cite{yang2015efficient}. We define and implement seven \textit{user-preferred} refinement operations for this approach. 

\subsection{Refinement Implementations}
\label{sec:ref_impl}
Our three systems support the following seven refinements that users request in \abr{hltm} systems ~\cite{musialek2016using, lee2017human}:

\paragraph{Remove word} $w$ from topic $t$. For all three systems, first forget all $w$'s tokens $w_i$ from $t$. Then, for \textit{info-gibbs} and \textit{info-vb}, assign a very small prior\footnote{We use $\epsilon = 10^{-8}$}~$\epsilon$ to $w$ in $t$. For \textit{const-gibbs}, add a constraint\footnote{We use $log(\epsilon)$ to make it a soft constraint. Replacing it with -$\infty$ will make it a hard constraint.} $f_m(z, w, d)$, such that $f_m(z, w, d)=log(\epsilon)$ if $z=t$ and $w=x$, else assign $0$.

\paragraph{Add word} $w$ to topic $t$. For all three systems, first forget $w$ from all other topics. Then, for \textit{info-gibbs} and \textit{info-vb}, increase the prior of $w$ in $t$ by the difference between the topic-word counts of $w$ and topic\textquotesingle s top word $\hat{w}$ in $t$. For \textit{const-gibbs}, add a constraint $f_m(z, w, d)$, such that $f_m(z, w, d)$ = 0 if $z=t$ and $w=x$, else assign $log(\epsilon)$.

\paragraph{Remove document} $d$ from topic $t$. For all models, first forget the topic assignment for all words in the document~$d$. Then, for \textit{info-gibbs} and \textit{info-vb}, overwrite the previous prior value with a very small prior~$\epsilon$, to $t$ in $\alpha_d$. For \textit{const-gibbs}, add a constraint $f_m(z, w, d)$, such that $f_m(z, w, d) = log(\epsilon)$ if $z=t$ and $d=x$, else assign $0$.

\paragraph{Merge topics} $t_1$ and $t_2$ into a single topic, $t_1$. For \textit{info-gibbs} and \textit{const-gibbs}, assign $t_1$ to all tokens previously assigned to $t_2$. This effectively removes $t_2$ and updates $t_1$, which should represent both $t_1$ and $t_2$. For \textit{info-vb}, add counts from~$\lambda_{t_2}$ to~$\lambda_{t_1}$ and remove row from $\lambda$ corresponding to $t_2$.

\paragraph{Split topic} $t$ given seed words~$s$ into two topics, $t_n$, containing $s$, and $t$, without $s$. For each vocabulary word, move a fraction of probability mass from $t$ to $t_n$ as proposed by \cite{pleple2013interactive}. Then, for \textit{info-gibbs} and \textit{info-vb}, assign a high prior for all $s$ in $t_n$. Following \citeauthor{fan2017prior}, we use $100$ as the high prior. For \textit{const-gibbs}, to $s$ to $t_n$, add a constraint $f_m(z, w, d)$, such that $f_m(z, w, d) = 0$ if $z=t_n$ and $w=w_i \in s$, else assign $log(\epsilon)$.

\paragraph{Change word order}, such that $w_2$ is higher than $w_1$ in topic~$t$. 
In \textit{info-gibbs}, increase the prior of $w_2$ in $t$ by the topic word counts' difference~$n_{w_{1,t}}$ -$n_{w_{2,t}}$. In \textit{info-vb}, increase the prior by $\lambda_{t,w_1} - \lambda_{t,w_2}$. For \textit{const-gibbs}, compute the ratio~$r$ between the topic word counts' difference $n_{w_{1,t}} - n_{w_{2,t}}$ and the counts of word~$w_2$, which have any topic except~$t$, $n_{w_{2,x}, x \neq t}$. Then, add a constraint $f_m(z, w, d)$, such that $f_m(z, w, d) = 0$ if $z=t$ and $w=w_2$, else assign $\delta$ where $\delta = \log(\epsilon)$ if $r>1$ else $\delta = 1.0-r$.

\paragraph{Create topic} $t_n$, given seed words, $s$. First forget the topic assignment for all $s$. Then, for \textit{info-gibbs} and \textit{info-vb}, assign a high prior to $s$. For \textit{const-gibbs}, to assign $s$ to $t_n$, add a constraint $f_m(z, w, d)$, such that $f_m(z, w, d) = 0$ if $z=t_n$ and $w=w_i \in s$, else assign $log(\epsilon)$.

\section{Measuring Control}
\label{sec:control}

Prior work in interactive systems emphasizes the importance of doing what users ask, that is, \textit{end user control}~\cite{ shneiderman2010designing, amershi2014power}. However, \abr{hltm}, which must balance modeling the data well and fulfilling users' desires, can frustrate users when refinements are not applied as expected~\cite{smith2018closing}. Evaluation metrics such as topic coherence, perplexity, and log-likelihood measure how well topics model data, but are not sufficient to measure whether user feedback is incorporated as expected. Therefore, we propose new \textit{control} metrics to measure how well models reflect users' refinement intentions. 

Consider a topic, $t$, as a ranked word list sorted in descending order of their probabilities in $t$. Let $r_{w_t}^{M_1}$ denote the rank of a word $w$ in topic $t$ in model $M_1$. After applying a word-level refinement, the rank of $w$ in the updated model~$M_2$, is~$r_{w_t}^{M_2}$. For word-level refinements, such as \textbf{add word}, \textbf{remove word}, and \textbf{change word order}, compute \textit{control} as the ratio of the actual rank change, the absolute difference ($r_{w_t}^{M_1} - r_{w_t}^{M_2}$), and the expected rank change. A score of $1.0$ indicates that the model perfectly applied the refinement, while a negative score indicates the model did the opposite of what was desired. For \textbf{remove document}, use the same definition as \textbf{remove word} except consider a topic as a ranked document list. 

For \textbf{create topic}, compute \textit{control} as the ratio of the number of seed words in the created topic out of the total number of provided seed words. For \textbf{merge topics}, \textit{control} is defined as the ratio of the number of words in the merged topic which came from either of the parent topics, and the total number of words shown to a user. For \textbf{split topic}, \textit{control} is the average of the \textit{control} scores of parent topic and child topic, computed using the \textit{control} definition for \textbf{create topic}.

%% file: sections/30-results.tex
\begin{table*}[t]
\resizebox{\textwidth}{!}{%
\centering
\begin{tabular}{|l|l|l|l|l|l|l|l|l|l|}
\hline
 & \multicolumn{3}{l|}{\textit{const-gibbs}} & \multicolumn{3}{l|}{\textit{info-gibbs}} & \multicolumn{3}{l|}{\textit{info-vb}}  \\ 
 \hline
 & $C_{Rand}$ & $C_{Good}$ & $Q_{Good}$$^*$ & $C_{Rand}$ & $C_{Good}$  & $Q_{Good}$$^*$ & $C_{Rand}$ & $C_{Good}$  & $Q_{Good}$$^*$ \\
 \hline
 remove w & 1.0 (0.0) & 1.0 (0.0) & 5.4 (9.7) & 1.0 (0.0) & 1.0 (0.0) & 3.0 (8.9) & 1.0 (0.0) & 1.0, (0.0) & 1.2 (5.0) \\ \hline
 remove d & 1.0 (0.0) & 1.0 (0.0) & -1.7 (10.8) & 1.0 (0.0) & 1.0 (0.0) & .8 (4.5) & .72 (.4) & .85 (.25) & -6.0 (13.2) \\ \hline
 merge t & .97 (.05) & 1.0 (0.0) & 6.3 (8.7) & .96 (.05) & 1.0 (0.0) & -.43 (9.3) & .99 (.02) & .99 (.02) & 1.4 (9.8) \\ \hline
add w & .82 (.29) & .86 (.24) & 3.0 (9.4)  & 1.0 (0.0) & .98 (.03) & 3.1 (6.4) & .98 (.04) & .98 (.02) & 1.7 (5.6) \\ \hline
create t & .08 (.10) & .81 (.13) & -6.6 (13.7) & .98 (.11) & .98 (.04) & -11 (10.4) & 1.0 (0.0) & 1.0 (0.0) & -13.0 (8.4) \\ \hline
split t & .91 (.09) & .79 (.19) & 1.9 (17.9) & .93 (.06) &  .87 (.19) & -7.9 (13.5) & 1.0 (0.0) & .93 (.16) & -1.6 (8)\\ \hline
reorder w & .41 (.53) & .19 (.20) & 1.6 (7) & 1.19 (.46) & .56 (.24) & -1.0 (5.5) & 1.02 (.27) & .44 (.24) & -1.0 (5.1) \\ \hline
\end{tabular}}
\caption{Simulation results, reported as \textit{mean (SD)}: control with the random ($C_{Rand}$) and good ($C_{Good})$  users, and coherence deltas ($Q_{Good}$) for the good user (we omit coherence for the random user as the goal there is not to improve the topics).  $^{*}$values reported as E-04.} 
\label{table:all_results}
\end{table*}

\section{\abr{hltm} System Comparison}

To compare how the three \abr{hltm} systems model data and adhere to user feedback (i.e., provide control), we need user data; however, real user interaction is expensive to obtain. So, we simulate a range of user behavior with these systems: users that aim to improve topics, ``good users'', and those that behave unexpectedly, ``random users''. 

The simulations use a data set of $7000$ news articles, $500$ articles each for fourteen different news categories
, such as business, law, and money, collected using the Guardian \abr{api}.\footnote{https://open-platform.theguardian.com} 

\subsection{Simulated Users}
The ``random user'' refines randomly. For example, \textbf{remove document}, deletes a randomly selected document from a randomly selected topic.

Our ``good user'' reflects a realistic user behavior pattern: identify a mixed category topic and apply refinements to focus the topic on its most dominant category.
Thus the ``good user''---with access to true document categories---first chooses a topic associated with multiple categories of documents and determines the dominant category of the top documents for the topic. Then, refinement operations push the topic to the dominant category. For example, the ``good user'' may remove a document which does not belong to the dominant category. Additional simulation are found in Appendix~\ref{sec:app_sim}. 
 
\subsection{Method}
We train forty initial \abr{lda} models, twenty with ten topics and twenty with twenty topics for the news articles, resulting in models with less and more topics than the true number of categories. 

For each of the three \abr{hltm} systems and each of the seven refinement types, we randomly select one of the pre-trained models. The create and split topic refinement types select from the models with ten topics, ensuring that topics have overlapping categories, while the others select from the models with twenty topics. We then apply a refinement as dictated by the simulated user. For the ``random user'', we randomly select refinement parameters, such as topic and word (Appendix ~\ref{sec:app_rand}), and for the ``good user'', we choose topic and refinement parameters intending to improve the topics (Appendix ~\ref{sec:app_good}).  We apply the refinement (Section~\ref{sec:ref_impl}) and run inference until the model converges or reaches a threshold of twenty Gibbs sampling and three \abr{em} iterations. We compute control (Section~\ref{sec:control}) of the refinement and change in topic coherence using \abr{npmi} derived from Wikipedia for the top twenty topic words~\cite{lau2014machine}.
We repeat this process $100$ times for each refinement type, simulated user, and \abr{hltm} system.

\section{Informed Priors Listen to Users, while Constraints Produce Coherent Topics}

Table~\ref{table:all_results} shows the per-refinement control and coherence deltas for the three different \abr{hltm} systems. As detailed in Appendix~\ref{sec:kw_results}, Kruskal-Wallis tests show that \abr{hltm} systems have significantly different ($p < .05$) control scores for all refinements for the ``good user'' and for all but \textbf{remove word} for the ``random user.''
Coherence deltas were also significantly different for all refinements except \textbf{add word}, where \textit{const-gibbs} yields consistently higher coherence improvements than the other conditions aside from \textbf{remove document}. 

For \textbf{remove word}, and \textbf{merge topics}, all methods provide good control (scores close to $1.0$). However, the informed prior methods, \textit{info-vb} and \textit{info-gibbs}, provide more control, for both the random ($C_{Rand}$) and good ($C_{Good}$) users, compared to \textit{const-gibbs}. Informed prior methods also excel at refinements that promote topic words, such as \textbf{add word} and \textbf{create topic}. On the other hand, \textit{const-gibbs} supports defining token and document-level constraints, which ensure almost perfect control for refinements that require restricting certain words or documents, such as \textbf{remove word} and \textbf{remove document}.

Additionally, comparing good and random users, all systems provide similar control except for \textit{const-gibbs} for \textbf{create topic}: $.81$ for good ($C_{Good}$) compared to $.08$ for random ($C_{Rand}$). This is because \textit{const-gibbs} is limited by the underlying data and cannot generate topics containing random, unrelated seed words, lowering control for the ``random user.'' 
Informed prior models, however, inflate priors to adhere to user feedback, regardless of whether it aligns with the underlying data, so these methods provide higher control even for random input. Finally, for \textbf{change word order}, all three systems lack control. As topic models are probabilistic models, it is therefore difficult to maintain the exact user provided word order. 

\subsection{Why Informed Priors Offer Control}
Informed priors provide higher control than constraints for refinements that require promoting words, such as \textbf{add word} and \textbf{create topic}. To understand the difference between these two feedback techniques, we conduct an additional simulation to compare \textit{const-gibbs} and \textit{info-gibbs}: we generate an initial topic model of $10$ topics and apply \textbf{add word} refinements to explore varied control of the feedback techniques. 


The initial model includes a \underline{law} topic with the top ten words: ``court, law, justice, rights, legal, case, police, human, public, courts''. A user wants to add the word ``injustice'', initially ranked at $1035$\textsuperscript{th} position, to this topic using both \textit{const-gibbs} and \textit{info-gibbs} models. While \textit{const-gibbs} improves the ranking of the added word to $631$, \textit{info-gibbs} puts this word at the first position in the updated topic. The \textit{const-gibbs} system tries to push tokens of ``injustice'' to the \textit{law} topic; however, there just are not enough occurrences to put it in the first ten words. Even assigning all its occurrences to the \underline{law} topic cannot improve its ranking further. On the other hand, \textit{info-gibbs} can increase the prior for ``injustice'' enough to put the word in the top of the topic list; until overruled by data \textit{info-gibbs}, can use high priors to incorporate user feedback, resulting in higher control. 



%% file: sections/40-conclusion.tex
\section{Conclusion}
Informed prior models provide an effective way to incorporate different feedback into topic models, improving \textit{user control} and topic coherence, while constraints yield higher quality topics, but with less control. While we simulate user behavior for good and random users, future work should compare these systems with end users, as well as compare end user ratings of control with our proposed automated metrics.

Interactive models---by design---are balancing user insight with the truth of the data (and thus the world).  An important question for future models, especially interactive ones, is how to signal to the user when their desires do not comport with reality.  In such cases, control may not be a desired property of interactive systems.

%% file: sections/appendix.tex
\section{Simulation Details}
\label{sec:app_sim}
To simulate the behavior of the ``random user'' and ``good user'' for the three \abr{hltm} systems, we train $40$ initial LDA models, $20$ with $10$ topics and $20$ with $20$ topics for the news articles, resulting in models with less and more topics than the true number of categories. 

\subsection{Random User Simulation}
\label{sec:app_rand}
To simulate random user behavior, for each of the three systems and for each of the seven refinement types, we randomly select a pre-trained LDA model from the pool of models with $20$ topics. Then, we apply a refinement of that refinement type to the selected model. We randomly select refinement specific parameters, such as candidate topic, word to be added, and document to be deleted. We run inference until the model converges or reaches a limit. 
For Gibbs sampling models, \textit{info-gibbs} and \textit{const-gibbs}, we use $20$ iterations as limit and for the variational model, \textit{info-vb}, we use $3$ EM iterations as the limit. 
After applying the refinement, we compute the \textit{control} and coherence given the updated and initial model. We perform this $100$ times for each of the refinement types and \abr{hltm} systems. 

\subsection{Good User Simulation}
\label{sec:app_good}
For each category $c$ of the $14$ categories of the Guardian news dataset (art \& design, business, education, environment, fashion, film, football, law, money, music, politics, science, sports, technology), we compute the most important words in $c$, $S_c$, using a Logistic regression classifier. We use $S_c$ as a list of representative words for category $c$.

Given a labeled corpus, we randomly choose one of the pre-trained models. When applying create or split topic refinement types, we select from the models with $10$ topics, ensuring that topics have overlapping categories. While applying all other refinement types, we select from the models with $20$ topics. We then simulate good user behavior for each of the refinement types as follows:

\begin{enumerate}
    \item Add word: Randomly select a topic $t$ from those where the top $20$ documents are from more than one category. Then, find the corresponding labeled category $c$ by analyzing top $20$ documents in the selected category. To improve the topic coherence of $t$, add top ranked words (from one to five words) from $S_c$, which are not already in the top words of $t$. 
    
    \item Remove word: Randomly select a topic $t$ from those where top $20$ documents are from more than one category. Then, find the corresponding labeled category $c$ by analyzing top $20$ documents in the selected category. For selected topic $t$, remove words which are not part of $S_c$.
    
    \item Change word order: Randomly select a topic $t$ among all topics. Then, find the corresponding labeled category $c$ by analyzing top $20$ documents in the selected category. Then, find words between index $10$ to $20$, which are at higher rank in $S_c$. Promote such words to a higher rank using change word order.
    
    \item Remove document: Randomly select a topic $t$ from those where top $20$ documents are from more than one category. Then, find the corresponding labeled category $c$ by analyzing top $20$ documents in the selected category. For selected topic $t$, delete documents (from one to five documents), which are not in $c$.
    
    \item Merge topics: 
    Randomly choose a topic pair to merge which represents a common category $c$.
    
    \item Create topic: 
    Randomly select a category $c$ which is not a dominant category in any of the topics. Create a topic by providing top $10$ words as seed words from $S_c$.
    
    \item Split topic: 
    Randomly select a topic from those which have documents from two different categories, $c_1$ and $c_2$. Split the top $20$ words in that topic into two lists using the representative words from $S_{c_1}$ and $S_{c_2}$. Then, split the topic using one of the lists.  

\end{enumerate}

\section{Kruskal Wallis Tests}
\label{sec:kw_results}
We provide details on the Kruskal Wallis tests used to assess whether there are significant differences in how the three \abr{hltm} systems, \textit{const-gibbs}, \textit{info-gibbs}, and \textit{info-vb}, impact control and topic coherence. The means reported here repeat what is provided in the main paper, but with the additional $\chi^2$ and $p$ values output from the Kruskal Wallis tests; $p < .05$ is considered to be significant. 

Because control values are not comparable across the seven \textit{user-preferred} refinements, we conducted separate Kruskal Wallis tests for each refinement.
The results include control for the simulated good user (Table~\ref{table:good_control}) and for the simulated random user (Table~\ref{table:random_control}), as well as quality improvements (coherence) for the simulated good user (Table~\ref{table:good_user_coherence}). 

\begin{table}[h!]
\resizebox{.5\textwidth}{!}{%
\begin{tabular}{|l|l|l|l|l|l|}
\hline
 & \textit{const-gibbs} & \textit{info-gibbs} & \textit{info-vb} & $\chi^2$ & p-value \\ \hline
add w & 0.82 & 1.00 & 0.99 & 249.35 & $< .001$ \\ \hline
remove w & 1.00 & 1.00 & 1.00 & 0.42 & .810 \\ \hline
remove d & 1.00 & 1.00 & 0.72 & 27.12 & $< .001$ \\ \hline
merge t & 0.97 & 0.96 & 0.99 & 31.24 & $< .001$ \\ \hline
reorder w & 0.41 & 1.19 & 1.03 & 113.52 & $< .001$ \\ \hline
create t & 0.08 & 0.98 & 1.00 & 277.23 & $< .001$ \\ \hline
split t & 0.91 & 0.93 & 1.00 & 119.47 & $< .001$ \\ \hline
\end{tabular}
}
\caption{Average control provided by the three \abr{hltm} systems for seven \textit{user-preferred} refinements and simulated random user behavior. Kruskal-Wallis tests ($p < .05$) show significant differences between the systems for all refinements except remove word.}
\label{table:random_control}
\end{table}

\begin{table}[h!]
\resizebox{.5\textwidth}{!}{%
\begin{tabular}{|l|l|l|l|l|l|}
\hline
          & \textit{const-gibbs} & \textit{info-gibbs} & \textit{info-vb}  & $\chi^2$  & p-value     \\ \hline
add w           & 0.86    & 0.98   & 0.98 & 13.02 & .001 \\ \hline
remove w        & 0.99    & 0.99   & 0.99 & 6.22  & .045 \\ \hline
remove d    & 0.99    & 0.99    & 0.85 & 163.73 & $< .001$    \\ \hline
merge t        & 1.00           & 1.00          & 0.99   & 22.76 & $< .001$    \\ \hline
reorder w & 0.19    & 0.56   & 0.44 & 103.44 & $< .001$    \\ \hline
create t       & 0.82   & 0.98   & 1.00        & 191.82 & $< .001$    \\ \hline
split t        & 0.77    & 0.87   & 0.93 & 81.71 & $< .001$    \\ \hline
\end{tabular}
}
\caption{Average control provided by the three \abr{hltm} systems for seven \textit{user-preferred} refinements and simulated good user behavior. Kruskal-Wallis tests ($p < .05$) show significant differences between the systems for all refinements.}
\label{table:good_control}
\end{table}

\begin{table}[h!]
\resizebox{.5\textwidth}{!}{%
\begin{tabular}{|l|l|l|l|l|l|}
\hline
          & \textit{const-gibbs} & \textit{info-gibbs} & \textit{info-vb}  & $\chi^2$  & p-value     \\ \hline
add w           & 3.0E-04    & 3.1E-04   & 1.7E-04  & 2.93 & .230  \\ \hline
remove w        & 5.3E-04    & 3.0E-04   & 1.2E-04 & 25.51 & $< .001$    \\ \hline
remove d    & -1.7E-04   & 7.5E-05   & -6.0E-04 & 19.29 & $< .001$    \\ \hline
merge t        & 6.3E-04    & -4.3E-05  & 1.4E-04  & 30.66 & $< .001$    \\ \hline
reorder w & 1.6E-04    & -8.0E-05  & -1.0E-05 & 7.67 & .020   \\ \hline
create t       & -6.6E-04   & -1.1E-03  & -1.2E-03 & 11.20 & .004 \\ \hline
split t        & 1.9E-04   & -7.9E-04  & -1.6E-04 & 22.19 & $< .001$    \\ \hline
\end{tabular}
}
\caption{Average coherence provided by the three \abr{hltm} systems for seven \textit{user-preferred} refinements and simulated good user behavior. Kruskal-Wallis tests ($p < .05$) show significant differences between the systems for all refinements except for add word.}
\label{table:good_user_coherence}
\end{table}